%% file: main.tex
\documentclass[letterpaper, 10 pt, journal, twoside]{IEEEtran}
\usepackage{times}
\usepackage{amsmath,amsfonts}
\usepackage{array}
\usepackage{textcomp}
\usepackage{stfloats}
\usepackage{url}
\usepackage{verbatim}
\usepackage{graphicx}
\usepackage[font=footnotesize]{caption}
\usepackage{cite}
\usepackage[bookmarks]{hyperref}
\usepackage{multirow}

\usepackage{booktabs}  
\usepackage{xcolor}
\begin{document}

\title{
Reactive Motion Generation via \\ Phase-varying Neural Potential Functions
}
\author{Ahmet Tekden$^{1}$ \and Dimitrios Kanoulas$^{2}$ \and Aude Billard$^{3}$ \and Yasemin Bekiroglu$^{1,2}$
    \thanks{Manuscript received: November 26, 2025; Revised March 7, 2026; Accepted April 3, 2026.}
    \thanks{This paper was recommended for publication by Editor Jens Kober upon evaluation of the Associate Editor and Reviewers' comments.}
    \thanks{This work was supported by Chalmers AI Research Center (CHAIR) and Chalmers Gender Initiative for Excellence (Genie), and the Wallenberg AI, Autonomous Systems and Software Program (WASP). D. Kanoulas was supported by UKRI FLF [MR/V025333/1] (RoboHike). A. Billard was supported by the euROBIN (European Robotics and AI Network) project under grant agreement No. 101070596. $^{1}$Chalmers University of Technology, Sweden. $^{2}$University College London, UK. $^{3}$Ecole Polytechnique Federale de Lausanne (EPFL), Switzerland. Email: {\tt\footnotesize tekden@chalmers.se}}
    \thanks{Digital Object Identifier (DOI): see top of this page.}
}
\maketitle
\vspace{-2em}
\begin{abstract}
Dynamical systems (DS) methods for Learning-from-Demonstration (LfD) provide stable, continuous policies from few demonstrations. First-order dynamical systems (DS) are effective for many point-to-point and periodic tasks, as long as a unique velocity is defined for each state. For tasks with intersections (e.g., drawing an “8”), extensions such as second-order dynamics or phase variables are often used. However, by incorporating velocity, second-order models become sensitive to disturbances near intersections, as velocity is used to disambiguate motion direction. Moreover, this disambiguation may fail when nearly identical position–velocity pairs correspond to different onward motions. In contrast, phase-based methods rely on open-loop time or phase variables, which limit their ability to recover after perturbations. We introduce Phase-varying Neural Potential Functions (PNPF), an LfD framework that conditions a potential function on a phase variable which is estimated directly from state progression, rather than on open-loop temporal inputs. This phase variable allows the system to handle state revisits, while the learned potential function generates local vector fields for reactive and stable control. PNPF generalizes effectively across point-to-point, periodic, and full 6D motion tasks, outperforms existing baselines on trajectories with intersections, and demonstrates robust performance in real-time robotic manipulation under external disturbances.
\end{abstract}
\section{Introduction}

\IEEEPARstart{L}{}earning-from-Demonstration (LfD) infers task policies directly from demonstration data, removing the need for manual design. A central challenge in LFD is producing motions that follow demonstrations while remaining robust to perturbations. Among LfD approaches, dynamical systems (DS)-based methods represent motions using ordinary differential equations (ODEs)~\cite{dmp, khansari2011learning}, enabling stable motion generation from few demonstrations. Time- or phase-dependent DS approaches use an explicit temporal variable~\cite{dmp}, allowing modeling of tasks where the same state is revisited at different times (Figure~\ref{fig:first-page}). However, reliance on a temporal variable limits their ability to recover smoothly after perturbations. In contrast, time-independent DS approaches generate motions from the current state alone~\cite{khansari2011learning, figueroa2018physically, figueroa2020dynamical,  khadivar2021learning, perez2023stable, node-clf-cbm}, offering greater reactivity. For tasks with intersections, they can be extended with velocity information to model the required motions. However, this introduces two key limitations: (i) sensitivity to disturbances near intersections due to reliance on velocity to disambiguate motion direction, and (ii) ambiguity when similar motion segments yield nearly identical position–velocity pairs that require different onward actions (e.g., $\dot{x}_1$ and $\dot{x}_2$ in knotting task, Figure~\ref{fig:first-page}).

\begin{figure}[t]
  \centering
\includegraphics[clip,trim=0 0 0 13cm,width=0.85\linewidth]{Figures/first_page.png}
  \caption{
  Reactive motion generation for tasks (6D trajectories) where the robot revisits similar states: (A) point-to-point knotting, with identical motion segments branching into different actions $\dot{x}_1, \dot{x}_2$ from the same position–velocity state (ambiguous for second-order models), and (B) 8-shaped periodic wiping.
  } 
  \label{fig:first-page}
\end{figure} 

DS approaches face a trade-off between modeling capability and reactivity to perturbations. We address this with Phase-varying Neural Potential Functions (PNPF)~\footnote{Project page: \url{https://fzaero.github.io/PNPF/}}, artificial potential functions conditioned on a novel closed-loop phase variable. \textit{The phase is estimated directly from state progression}, ensuring that the DS remains time-invariant. PNPF combines two energy functions: (i) a nominal energy, which encodes the desired task behavior and decreases monotonically with task progress; and (ii) a safety energy, which discourages leaving the boundary of demonstrated data, with this boundary estimated in a data-driven manner from the visited states.

The nominal energy provides a scalar measure of task progress, enabling closed-loop estimation of the phase variable directly from the state without relying on time or velocity. This addresses the two main weaknesses of second-order dynamical system models: their sensitivity to velocity disturbances and the ambiguity introduced by repeated states. 

The safety energy guides the system back toward the region validated by demonstrations whenever it moves outside, while still allowing compliant behavior within. This region encodes the workspace and task constraints implicitly followed by the demonstrator, representing the part of the state space where reliable task execution has been observed. Staying within this region is desirable because it ensures that the robot operates in areas well-supported by data, minimizing the risk of unsafe or infeasible behavior. At the same time, the system can adapt to variations within the demonstrated region. For example, if the robot is subjected to an external disturbance that moves it to a nearby state within the demonstrated workspace, or if it encounters obstacles, it can adjust its trajectory while remaining in the reliable region. This approach balances the need for safety and reliability while allowing flexibility within the validated region. 

This modular design of PNPF separates phase estimation from demonstration-based safety constraints and enables the integration of additional energy terms, such as those for obstacle avoidance, as demonstrated in the experiments. The method handles point-to-point and periodic motions, generalizes to tasks in configuration or task space, and supports full 6D motion while preserving reactivity.

Our main contributions are: (i) a reactive motion generation framework based on neural potential functions that models point-to-point and periodic motions from few demonstrations, while incorporating a data-driven safety set, to keep the robot within states observed in demonstrations; (ii) a closed-loop phase variable that tracks task progress while preserving reactivity to external disturbances; and (iii) compatibility with configuration- and task-space motion generation, including full 6D tasks. We evaluate our approach on 2D handwriting and 6D real-robot tasks, benchmarking it against state-of-the-art reactive motion generation methods. PNPF enables robust real-robot execution and outperforms baselines in overall performance, particularly in scenarios with intersecting trajectories where other methods fail.

\vspace{-1em}
\section{Related work}
LfD is a widely adopted paradigm in robot learning for acquiring new skills~\cite{argall2009survey, schaal1996learning}. Prominent approaches include state-only DS models~\cite{khansari2011learning,  khansari2017learning, figueroa2018physically,  rana2020learning, figueroa2020dynamical, bahl2020neural, khadivar2021learning, perez2023stable, node-clf-cbm, totsila2025sensorimotor}, and time- or phase-dependent motion generation methods, such as movement primitives~\cite{dmp, promp, cnmp, li2023prodmp, tekden2023neural, yildirim2024conditional}. State-only DS models represent motion as vector fields, enabling online reactivity to perturbations~\cite{billard2022learning}. To handle tasks with intersecting trajectories, second-order representations incorporating velocity are preferred. However, these models can be sensitive to disturbances near intersections and may become ambiguous when intersections are preceded by similar motion segments (Sec.~I). In contrast, movement primitives manage intersections using time or phase variables, but this comes at the cost of reduced reactivity, since progression is governed by an open-loop schedule rather than the current state. Our approach combines the strengths of both families by using a closed-loop phase variable to parameterize a potential function, preserving DS-style reactivity and matching the modeling capacity of movement primitives for point-to-point and periodic tasks.

\begin{figure}[!t]
    \centering
    \includegraphics[width=0.80\linewidth]{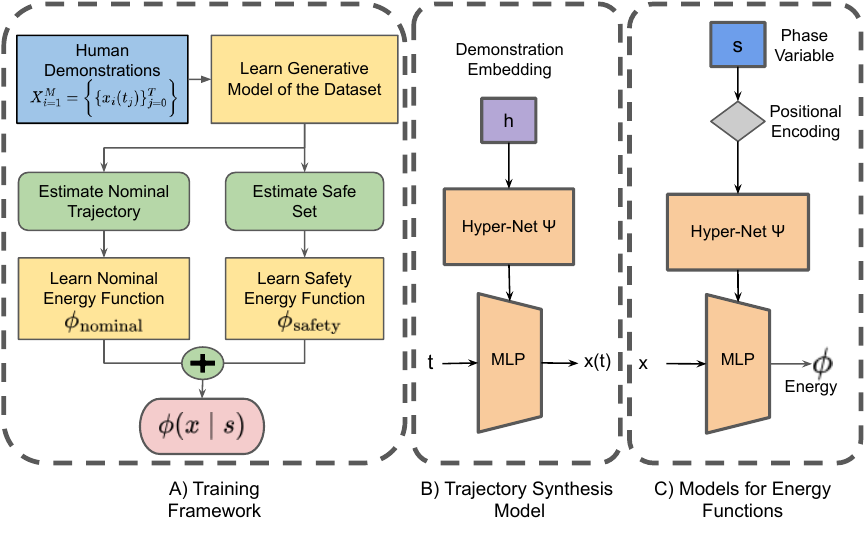}
    \caption{The proposed model architecture: (A) shows how PNPFs $\phi(x\mid s)$ are constructed. (B) and (C) shows the neural network architectures utilized for trajectory generation and energy function estimation. 
    }
    \label{fig:framework-models}
\end{figure}

Extensions of DS models include discrete sequencing and locally active regions. In the former, complex skills are divided into subtasks, each modeled as a stable DS, with sequencing handled by a Hidden Markov Model (HMM)~\cite{medina2017learning}. The HMM introduces a discrete notion of progress: subtask switching is governed by transition probabilities that increase near the attractor of the current mode. This allows revisiting the same state across different stages of the task, but subtask switching can be delayed under perturbations, as transitions depend on attractor proximity. In the latter, locally active regions impose stiffness-like attraction around a reference trajectory while preserving global stability~\cite{figueroa2022locally}. These regions are predefined rather than derived from demonstrations and tie recovery to a single path. In contrast, our method provides continuous progress via a closed-loop phase variable estimated directly from state progression, and learns attraction to demonstration regions obtained from data, enabling robust sequencing and safe recovery. Regions analogous to our safety boundary can also be specified through control barrier functions (CBF)~\cite{ames2017cbf}. For instance, \cite{saveriano2019learning} constructs data-driven CBF safety constraints using a small set of linear hyperplanes defined by heuristic rules. This formulation, however, limits the safety region to coarse, convex, piecewise-linear shapes. Although convex curved sets can be approximated with additional hyperplanes, the approach does not readily scale to more complex geometries. In contrast, our safety energy models the boundary as a single smooth function, enabling a much wider variety of smooth shapes without relying on numerous local linear approximations.

Artificial potential functions~\cite{khatib1986real} generate reactive motion by shaping landscapes that attract the robot to targets and repel it from obstacles. Recent work has learned such functions from demonstrations~\cite{khansari2017learning,rana2020learning}, but often suffers from stiff dynamics~\cite{khansari2017learning} or underconstrained optimization~\cite{rana2020learning}, leading to unpredictable behavior in unseen states. Relying solely on state information also limits the modeling of tasks with repeated state visits. Our approach instead builds potential functions from synthesized trajectories, jointly encoding task behavior and safety regions, and conditions them on a closed-loop phase variable, enabling predictable behavior in unseen states and handling repeated state visits.

Smooth potential functions are crucial for generating continuous vector fields that support stable and reactive control. We achieve this using neural fields~\cite{nerf_survey}, coordinate-based networks that compactly represent continuous functions and have seen success in visual~\cite{deepsdf,siren}, and robotic~\cite{irshad2024neural, grasp_df1, tekden2023grasp,chen2025implicit, tekden2023neural} applications. Previous works have used neural fields for motion modeling~\cite{tekden2023neural}, but these methods were not designed for online use and lack the smoothness required for reactive control. We address this with a positional encoding tailored for smoothness, ensuring that the gradients of the energy function, i.e., the vector field, vary smoothly with the robot state, enabling stable and reactive real-time control.

In summary, we present an LfD framework that combines the reactivity of state-based DS with the expressiveness of movement primitives, outperforming prior approaches on complex motions and supporting real-time control.

\section{Motion generation with Potential functions}

In the proposed model, trajectories follow the gradient of a scalar-valued potential function, which defines an energy landscape over the state space that encodes goal-directed behavior. By following the gradient of the landscape, the desired motion arises naturally as trajectories that converge to low-energy regions representing task goals. Formally, we define a potential function $\phi(x \mid s)$ where $x \in \mathbb{R}^d$ is the state in the configuration or task space, and $s$ is a phase variable, denoting task progress. The policy is then expressed as $\dot{x} = - \nabla_x\phi(x\mid s)$. This formulation enables time-independent control while allowing tasks to revisit the same state at different timesteps via phase conditioning. Figure~\ref{fig:framework-models} illustrates the potential function construction process and the proposed neural network architectures.

To construct the potential function $\phi(x \mid s)$, we propose a design composed of two components: nominal and safety energy functions. The nominal energy function encodes task-specific behavior by guiding the system along an unknown nominal trajectory $x^*(t)$. It supports closed-loop estimation of the phase variable $s$ to keep the system synchronized with task progression (Sec.~\ref{sec:dynamic-potential-function}). The safety energy enforces active corrections by attracting the system toward a demonstration-based safety region, defined as a data-driven support region around the demonstration states and formally denoted as the safety set $C$. Given $M$ human demonstrations as state trajectories $X_{i=1}^M = \{x_i(t_0),x_i(t_1),...,x_i(t_T)\}$, we utilize a trajectory generation model to infer both the unknown nominal trajectory $x^*(t)$ and the safe set $C$. This structure yields motions that are both safe and consistent with demonstrated behavior.

\begin{figure*}[t]
    \centering
    \includegraphics[width=0.75\linewidth]{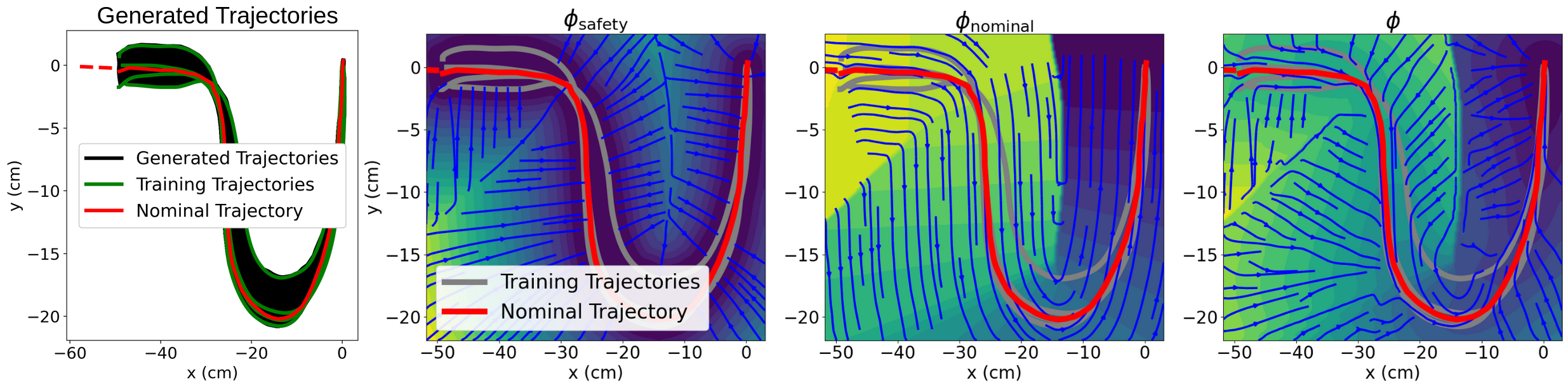}
    \caption{Generated trajectories, and the corresponding potential function in the form of $\phi(x)$. The generated trajectories densely cover the safe set, which contains the training demonstrations (green). The nominal trajectory (solid red) has the lowest DTW distance to demonstrations; the dashed segment shows its extrapolation.  $\phi_{\text{nominal}}$ guides the robot to follow the nominal behavior while $\phi_{\text{safety}}$ guides the robot back to the safe set.}
    \label{fig:energy-function-static}
\end{figure*}
\vspace{-1em}
\subsection{Trajectory synthesis through generative modeling}
We learn a generative model that produces trajectories similar to the demonstrations while densely filling the gaps between them. The training trajectories are time-aligned and parameterized over $t\in[0,T]$, and examples are shown in Figure~\ref{fig:energy-function-static}. These generated trajectories are used to estimate the nominal trajectory $x^*(t)$ and the safe set $C$.

We use decoder-only neural networks~\cite{deepsdf}, in which conditioning is performed via a hypernetwork~\cite{hypernet} to train the trajectory synthesis model. These architectures have fewer parameters to learn and work robustly even with a small number of data samples~\cite{tekden2023neural}. In decoder-only neural networks, instead of training an encoder to predict latent vectors from demonstrations, we directly assign a learnable embedding to each demonstration, which is optimized jointly with the network weights during training via backpropagation.

At inference, we estimate the nominal trajectory and the safe set by generating novel trajectories from latent vectors $h'$, sampled from the convex hull of the training embedding vectors $h_i$. This sampling procedure ensures that the generated trajectories remain within the interpolation region defined by the human demonstrations. We utilize the trajectory generation model to approximate the nominal trajectory $x^*(t)$ by selecting, among the generated trajectories $\{\tilde{x}_n(t)\}$, the one that minimizes the sum of dynamic time warping (DTW)~\cite{dtw} distances to the demonstrations, i.e., $x^*(t) = \arg\min_{\tilde{x}_n(t)} \sum_{i=1}^M \mathrm{DTW}(\tilde{x}_n(t), x_i(t))$. In addition, the beginning of the $x^*(t)$ is extended using a first-order polynomial to ensure the continuity of nominal energy at $t_1$.

We formally define the safety set, i.e., the safety region inferred from demonstrations, as  $\mathcal{C} = \{x \mid \text{SDF}(x) \leq 0\}$, where the continuous signed distance function (SDF), obtained from demonstration data, quantifies the distance to the boundary with negative values inside the region and positive values outside. We employ our trajectory model to generate additional samples, thereby enhancing the density of the safety region and the precision of boundary approximation. Specifically, we generate a large number of trajectories and label the states visited along these trajectories as belonging to the safety region, thereby constructing the set $X_{\text{in}}$. Points outside the safety region, denoted $X_{\text{out}}$ are obtained by sampling the state space and selecting those farther than a threshold $\epsilon_{\text{SDF}}$ from $X_{\text{in}}$. The SDF is then defined as the distance to the closest point in the opposite set: $\text{SDF}(x) = \min_{x_{\text{in}} \in X_{\text{in}}} \lVert x - x_{\text{in}}\rVert$ if $x \in X_{\text{out}}$, and $\text{SDF}(x) = -\min_{x_{\text{out}} \in X_{\text{out}}} \lVert x - x_{\text{out}}\rVert$ if $x \in X_{\text{in}}$.

\vspace{-1em}
\subsection{Modeling energy functions for task representation}

Human demonstrations specify correct behavior only in the regions they visit, i.e., the safe set. Accordingly, we assume the robot can reliably complete the task by either (i) following the nominal trajectory within the safe set or (ii) returning to the safe set when outside.

Consequently, we derive two energy functions from the demonstrations: a safety energy function $\phi_{\text{safety}}$, which guides the robot back to the safe set, and a nominal energy function $\phi_{\text{nominal}}$, which captures the nominal behavior within the safe set. We first derive these functions based solely on the state $x$. Then, in the following subsection, we show how nominal energy is used as a phase variable.

\begin{figure*}[t]
    \centering
    \includegraphics[width=0.81\linewidth]{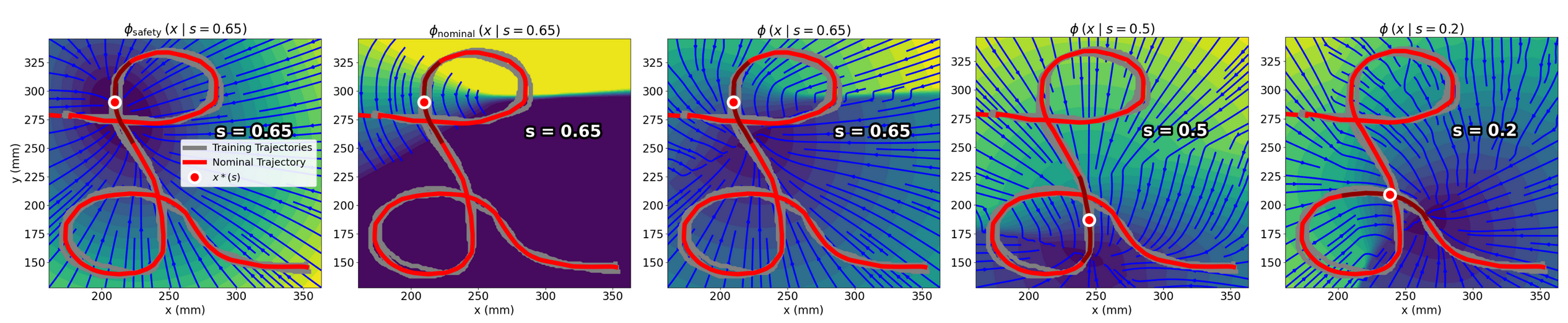}
    \caption{Example visualizations of $\phi_{\text{safety}}(x\mid s)$ and $\phi_{\text{nominal}}(x\mid s)$ used for estimating $\phi(x\mid s)$, along with $\phi(x\mid s)$ at different values of $s$. $s$ is normalized to be between $0$ and $1$.}
    \label{fig:potential-function-dynamic}
\end{figure*}

The safety energy is represented using the SDF estimated in the previous section and is given by $\phi_{\text{safety}}(x) = 
\text{relu}(\lambda_\text{safety}(\text{SDF}(x)))$ where $\lambda$ smooths the function near $0$, ensuring that the gradients of $\phi_{\text{safety}}$ decay gradually as $\text{SDF}(x)$ approaches $0$. By construction, $\phi_{\text{safety}}(x)\geq 0$ and vanishes for all $x \in \mathcal{C}$.

To define the nominal energy function, we use the remaining arc length of the nominal trajectory. Given the discretized nominal trajectory $\{x^*_j=x^*(t_j)\}_{j=0}^{N}$, where $N$ denotes the final index of the trajectory, the nominal energy at point $x^*_k$ is defined as $\phi_{\text{nominal}}(x^*_k) \;=\; \sum_{j=k}^{N-1} \lVert x^*_{j+1} - x^*_j\rVert$. This scalar decreases monotonically as the robot progresses along the trajectory and satisfies $\phi_{\text{nominal}}(x^*_N) = 0$. For states not on the nominal path, we approximate the energy by projecting the state onto the closest point on $x^*_k$, resulting in
\begin{equation}    
\phi_{\text{nominal}}(x) = \phi_{\text{nominal}}(x^*(\arg\min_k \lVert x^*_k - x\rVert)),
\end{equation}
which encourages motion parallel to the nominal trajectory and ensures $\phi_{\text{nominal}}(x)\geq 0$ for all $x$. The final potential function $\phi$ is expressed as the summation of safety and nominal energy functions: $\phi(x) = \phi_{\text{nominal}}(x) + k_{\text{safety}} \times \phi_{\text{safety}}(x)$ where $k_{\text{safety}}>0$ weights the safety term ($k_{\text{safety}}=1$ in all experiments). By construction, $\phi(x)\geq 0$ and becomes $0$ only at states inside the safety set where the nominal energy reaches zero, i.e., data-driven region corresponding to the terminal states of the demonstrated trajectories. This is illustrated in the example potential function shown in Figure~\ref{fig:energy-function-static}.
\vspace{-1em}
\subsection{Phase-varying potential function design}
\label{sec:dynamic-potential-function}

The potential function $\phi$ described in the previous section models many tasks effectively, but it cannot represent behaviors that revisit the same state at different timesteps. Since $\phi$ assigns a single scalar value to each state, it cannot encode multiple energy levels at the same location, making revisits impossible under gradient flow. To address this, we represent the potential function with a new phase variable derived from the nominal energy.

As the robot follows the nominal trajectory $x^*(t)$, the nominal energy decreases monotonically to 0. We can therefore reparameterize $x^*_k$ as $x^*(s_k)$, where $s_k = \phi_{\text{nominal}}(x^*_k)$. Using $s$ as a phase variable, we represent $\phi$ dynamically by limiting the estimation of the nominal and safety energy functions to the region around the current $s_t$ value~\footnote{In practice, $s$ is normalized to be between $1$ and $0$, by dividing it by $s_{0}$.~\footnotemark[2]\,We use $\Lambda(s)=1+(1-\tilde{s})^{10}$ with $\tilde{s}= s/s_0$ in all experiments.}. Specifically, given the energy window $s_w$, we estimate $\text{SDF}(x\mid s_k)$ and $\phi_{\text{safety}}(x\mid s_k)$ using the points $x\in X(s_k:s_k+s_w)$, and $\phi_{\text{nominal}}(x\mid s_k)$ based on the segment of the nominal trajectory $x^*(s_k-s_w:s_k+s_w)$. With these updates, the PNPF is expressed as:
\begin{equation}
    \phi(x\mid s) = \phi_{\text{nominal}}(x\mid s) + k_{\text{safety}} \; \Lambda_{\text{safety}}(s) \; \phi_{\text{safety}}(x\mid s)
\end{equation}
where $\Lambda_{\text{safety}}(s)$ increases the strength of the safety term as $s$ decreases~\footnotemark[2]. While not required, it biases the potential toward stronger safety convergence as the task progresses. Figure~\ref{fig:potential-function-dynamic} illustrates an example composition of the energy functions and potential functions at different $s$ values, showing how the PNPF model outputs different velocities at similar states. With these formulations, given the initial state $x$ and the phase variable $s$, the following equations are used to estimate the control signal and update the phase variable:

\setcounter{footnote}{2}

\begin{equation}
    \dot x = - \alpha \nabla_x\phi(x\mid s), \quad
    s \leftarrow \phi_\text{nominal}(x'\mid s)
    \label{s-equation}
\end{equation}
where $\alpha>0$ is the control gain and $x'$  is the new state after applying $\dot x$. This controller enables the robot to perform the task correctly while providing robustness to perturbations due to the closed-loop nature of the phase update.

Due to the compositional nature of energy functions, we can integrate additional terms into $\phi$, such as for obstacle avoidance or goal reaching. Following the standard artificial potential formulation~\cite{khatib1986real}, we represent goals as attractive potentials and obstacles as repulsive potentials. When the phase variable $s$ falls below a given threshold, the attractive potential term is enabled to ensure asymptotic convergence to the goal, while the repulsive term remains active throughout to maintain obstacle clearance.

\vspace{-1em}
\subsection{Stability of PNPF}
We consider the closed-loop system $\dot x = -\alpha \nabla_x \phi(x\mid s)$ with phase update $s \leftarrow \phi_{\text{nominal}}(x\mid s)$, and use $V(x,s) = \phi(x\mid s)$ as a Lyapunov candidate. The potential is the sum of nominal and safety energies, is nonnegative, and vanishes only on the target set $S$ given by the intersection of the safety set and the zero level set of $\phi_{\text{nominal}}$. Both terms are parameterized with ReLU neural fields, yielding continuous, piecewise-smooth energy functions (see Sec.~\ref{sec:nf}) with gradients well-defined almost everywhere, except on a measure-zero set where activations switch~\footnote{At these non-differentiable points, automatic differentiation returns a valid subgradient, ensuring the controller remains numerically well-defined. This is consistent with our piecewise formulation based on phase-windowed energies and a clipped safety term.}, inducing a locally Lipschitz closed-loop vector field in the data-supported region~\footnote{The data-supported region is the subset of the state space on which the model is trained. For each phase value s, training samples are generated as $x^*(s)+u$, where $u$ is drawn from a bounded Euclidean ball.}. Since summation may introduce spurious stationary points via gradient cancellation, we assume the demonstrations are locally consistent, i.e., $\nabla_x \phi_{\text{nominal}}(x\mid s)^\top \nabla_x \phi_{\text{safety}}(x\mid s) \ge -\varepsilon||\nabla_x \phi_{\text{nominal}}(x\mid s)||^2$, for small $\varepsilon>0$, which preserves a descent direction for $V$, since $\dot{V} = -\alpha \|\nabla_x \phi\|^2$ and $\|\nabla_x \phi\|^2 \ge \|\nabla_x \phi_{\text{safety}}\|^2 + (1-2\varepsilon)\|\nabla_x \phi_{\text{nominal}}\|^2 > 0.$ This assumption limits the method to unimodal tasks, where aligned velocities at each phase yield locally consistent motion directions. The phase update is discrete (see Eq.~(7)); for analysis, we use a local continuous-time approximation valid for $|\Delta s| < s_w$, under which the energy landscape can be treated as locally invariant in $s$, yielding $ \dot s = \nabla_x \phi_{\text{nominal}}(x\mid s)^\top \dot x = -\alpha \nabla_x \phi_{\text{nominal}}(x\mid s)^\top \nabla_x \phi(x\mid s)$. Under the above assumption $\nabla_x \phi_{\text{nominal}}(x\mid s)^\top \nabla_x \phi(x\mid s) \ge (1-\varepsilon) \|\nabla_x \phi_{\text{nominal}}(x\mid s)\|^2 >0$, and therefore $\dot s < 0$ for $s_0 > s(0) > 0$, implying monotonic phase decrease and $s(t) \to 0$. Along trajectories, $\dot V \le 0$, and by LaSalle’s invariance principle, trajectories initialized in the data-supported region converge to the largest invariant subset of $S$, establishing regional asymptotic convergence. 

\vspace{-1em}
\subsection{Modeling smooth potential functions with neural fields}\label{sec:nf}

We model $\phi_{\text{safety}}(x\mid s)$ and $\phi_{\text{nominal}}(x\mid s)$ using neural fields, as they are well suited to represent continuous functions and support conditional modeling, including with phase variables. These networks are fully differentiable, allowing us to estimate $\dot x$ as $-\nabla_x\phi(x\mid s)$. Maintaining continuous gradients in the potential function supports smooth robotic motion, as discontinuities or plateaus can lead to getting stuck in local minima. 

Positional encoding~\cite{tancik2020fourier} is commonly used in neural fields to represent complex functions but may introduce gradient discontinuities~\cite{siren}. To overcome this, we use a hyper-network-conditioned MLP, where the hyper-network takes the position-encoded phase variable $s$, and the MLP takes the non-position-encoded state variable $x$. This design leverages the known spectral bias of standard MLPs with ReLU activations~\cite{tancik2020fourier}, which tend to represent smoother, low-frequency functions when operating directly on coordinates, and we harness this property to produce smooth energy landscapes in $x$. This design captures motion patterns with sharp curvature changes and loops over time. At the same time, it preserves locally smooth energy landscapes in $x$ by modeling phase-windowed regions using non-position-encoded state inputs. Training uses states sampled from neighborhoods of $x^*(s)$ at each phase. However, since the energy functions are learned, we additionally include a lightweight sampling-based safeguard that activates only when gradient-based progress stalls. The safeguard uses the potential as a running cost by sampling candidate actions and evaluating them via one-step rollouts. We train all energy functions using the AdamW~\cite{loshchilovdecoupled} optimizer. 

\vspace{-1em}
\subsection{Modeling periodic motions}
To model periodic motions, we modify the input of the trajectory generation model and the energy functions to be periodic. Given period $\mathcal{T}$ corresponding to one cycle, the trajectory generation model then takes input $ t^\circ = \left[\sin\left(2\pi \frac{t}{\mathcal{T}}\right), \cos\left(2\pi \frac{t}{\mathcal{T}}\right)\right],$ and the energy functions take phase input $s_k^\circ = \big[\sin(2\pi \frac{s_k}{s_\mathcal{T}}), \cos(2\pi \frac{s_k}{s_\mathcal{T}})\big]$, where $s_{\mathcal{T}}$ denotes the arc length of one nominal period. Accordingly, $\phi$ is now expressed as:
\begin{equation}
\phi(x \mid s) =
\begin{cases}
\phi_{\text{nominal}}(x \mid s^\circ) + k_{\text{safety}} \, \phi_{\text{safety}}(x \mid s^\circ) &\hspace{-0.5em}\text{if } s \ge 0,\\
k_{\text{safety}} \, \phi_{\text{safety}}(x \mid s^\circ) & \hspace{-1.2em}\text{otherwise}.
\end{cases}
\end{equation}
When $s\le0$, the second condition will lead the motion to conclude while still providing a stable DS equation. With this new formulation, we use the following phase update:
\begin{equation}
s \leftarrow s - \left(\phi_{\text{nominal}}(x \mid s^\circ) - \phi_{\text{nominal}}(x' \mid s^\circ)\right).
\end{equation}

\vspace{-1.1em}
\subsection{Modeling 6D motions}
For motions with both position and orientation (unit quaternions), we follow~\cite{ude_quat,clfd} and represent orientations in axis-angle form. Given a quaternion trajectory $q_{1:T}$, we express the axis-angle trajectory $r_{1:T}$ in the tangent space of the final quaternion $q_T$ (for periodic motions, $q_1$ is used instead) as $r_t = \mathrm{Log}(q_t * \bar q_T)$, where $\bar q_T$ is the conjugate of $q_T$, $*$ denotes quaternion multiplication, and Log is the logarithmic map from quaternions to axis-angle vectors~\cite{explog_map}. The resulting $r_{1:T}$ is continuous and can be modeled in the same way as Cartesian trajectories. At inference time, $-\nabla\phi$ yields angular velocity for direct robot control, or predicted $\hat r_{1:T}$ can be mapped back via $\hat q_t = \mathrm{Exp}(\hat r_t) * q_T$, with Exp denoting the exponential map from axis-angle to quaternions~\cite{explog_map}.

\vspace{-1em}
\section{Simulation-based evaluations} 

In this section, we first evaluate the performance of our model on three state-of-the-art simulated datasets: LASA~\cite{khansari2011learning} (2D), LAIR~\cite{perez2023stable} (2D), RoboTasks~\cite{clfd, clfd-snode} (6D). We also introduce two new 2D datasets, CHAR and CHAR-Periodic, both of which are more challenging than LAIR~\footnote{We limit the evaluation to unimodal tasks.}. In CHAR, trajectory intersections are preceded by similar motion segments but require different onward action predictions. As a result, while second-order state representations are sufficient for LAIR, they fail on CHAR. We also demonstrate how our model can react to obstacles and spatial disturbances without affecting the geometry of the motion. For comparative evaluation, we benchmark against two baseline models: NODE~\cite{node-clf-cbm} and CONDOR~\cite{perez2023stable}~\footnote{The original implementation for CONDOR uses the same trajectories for training and testing; we modified it to separate them, leading to slightly lower performance than originally reported.}. Note that CONDOR natively supports second-order state representations.

\input{Results/short_results_one_column}

Most reactive motion generation methods for periodic motion modeling rely on first-order state-only DS formulations, which are insufficient for the CHAR-Periodic dataset. For experiments with this dataset, we introduce a naive baseline that repeats the first periodic segment of the training trajectory three times to match the test length.

\begin{figure}[t]
    \centering
    \includegraphics[width=0.84\linewidth]{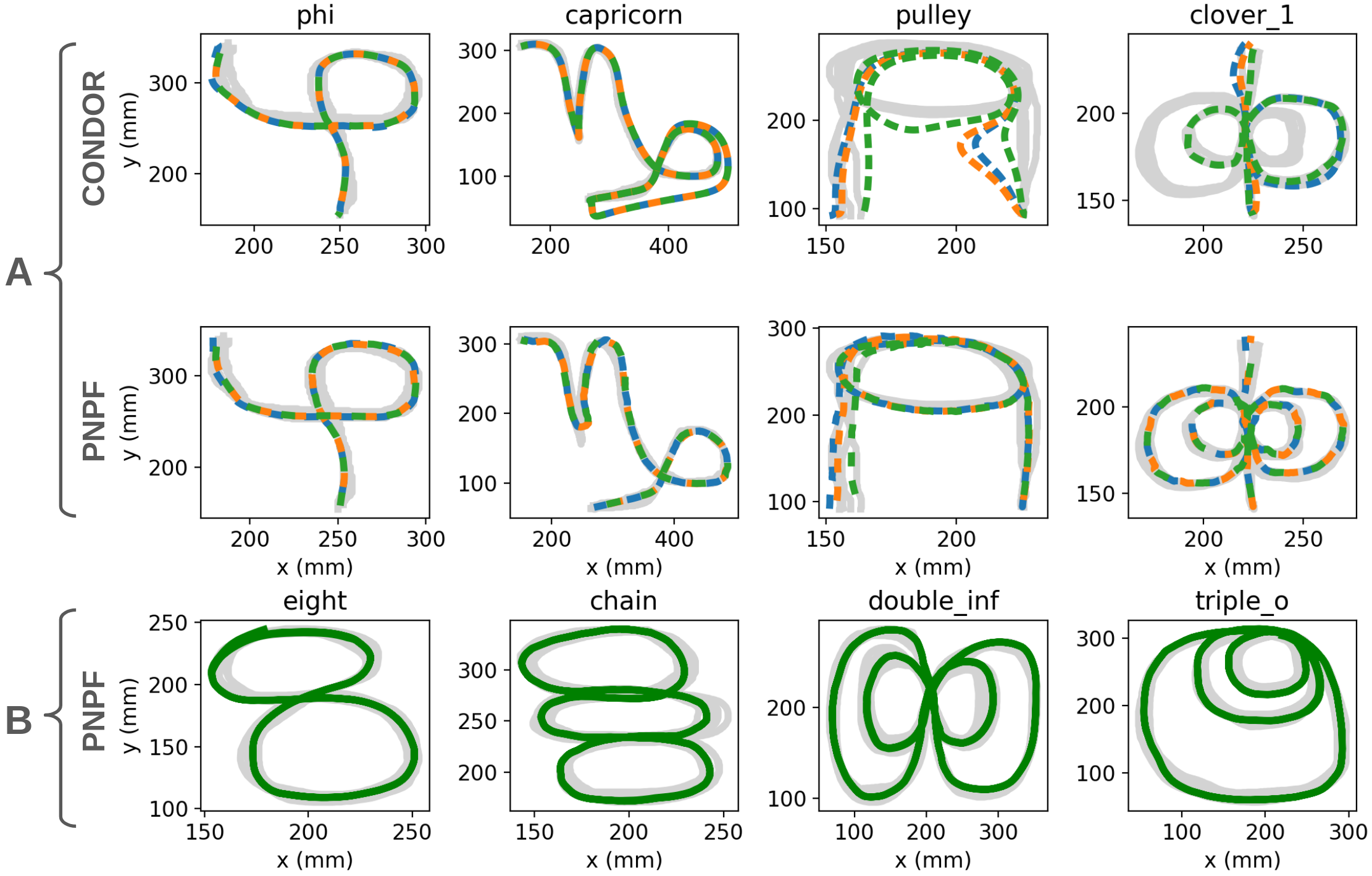}
    \caption{Comparison with CONDOR (A), and periodic motions from our method (B). Grey lines are demonstrations; dashed colored lines (A) are reproductions from different initial states. PNPF reproduces motions more faithfully, while CONDOR mismodels (capricorn) and misses segments (pulley, clover\_1). In capricorn, our method smooths out sharp corners.
    }
    \label{fig:condor-pnpf-comp}
\end{figure}

For the 2D motion datasets, we evaluate the performance of the model using four metrics: DTW Distance (DTWD), Fréchet Distance (FD)~\cite{eiter1994computing}, Final Position (FP) error, and accuracy. To measure the accuracy, we define a prediction as successful if the distance between the final point of the ground truth and the predicted trajectory is within 2 cm. The FP error is omitted for periodic motions, as these typically lack a well-defined endpoint. For RoboTasks, we use the DTWD for positions (DTWD-P) and orientations (DTWD-O), FP error, and final orientation (FO) error as evaluation metrics. Among these metrics, DTWD and FD measure geometric fidelity, while FP error, FO error, and accuracy measure the stability of the predicted trajectories. To ensure consistency, all predicted trajectories are rescaled to length 1000 and DTWD is normalized accordingly. In addition, to improve the reliability of performance estimates and assess robustness to random initialization, all models are trained and evaluated using three different random seeds. 

The overall mean performance across all datasets~\footnote{NODE results are omitted for LAIR and CHAR, as the method cannot model tasks with repeated states.} is reported in Table~\ref{tab:dataset-comparison}, where our method achieves the highest average score across all baselines. In particular, in the CHAR dataset, which contains motions with repeated state visits at different timesteps, our method reduced DTWD by around $80\%$ compared to the baseline. We additionally evaluate the robustness of our method on this dataset to suboptimal nominal trajectory selection by selecting the candidate (i) with the highest DTWD (PNPF-H) and (ii) from a reduced pool of 50 trajectories (PNPF-50). While both settings degrade performance compared to the default configuration, they still achieve better results than the baseline across all metrics.\footnote{Note that artifacts in human demonstrations (e.g., knots) can propagate to the selected nominal trajectory and degrade performance.} 
\input{Results/Lair_results}
\begin{figure}[t]
    \centering
    \includegraphics[width=0.88\linewidth]{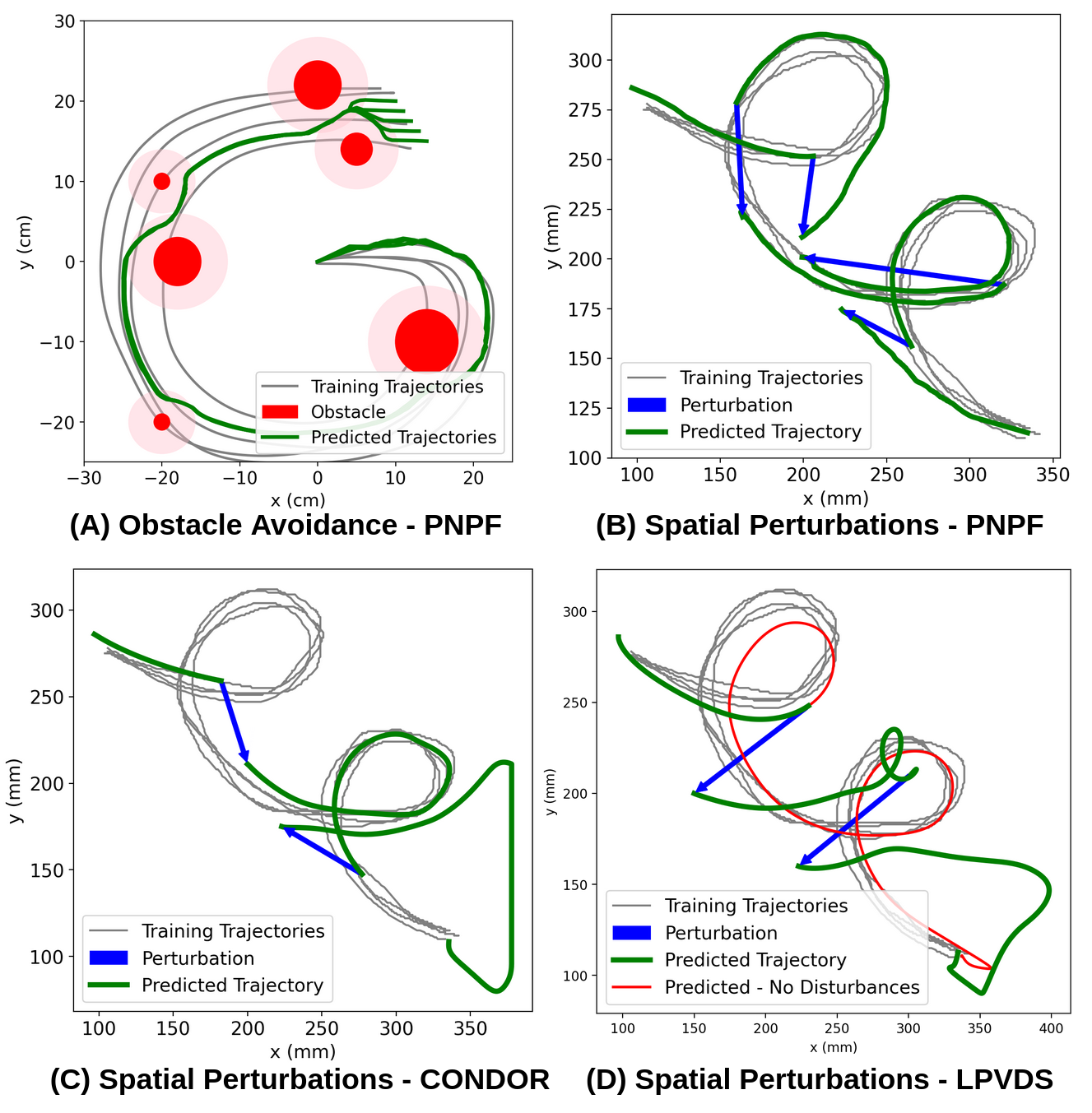}
    \caption{PNPF results for obstacle avoidance (A) and perturbations (B). CONDOR skips or repeats motion segments after perturbations (C), while LPVDS struggles to recover and drifts from the demonstrated path (D).}
    \label{fig:spatial-perturbations}
\end{figure}
Table~\ref{Lair-results} presents individual results for LAIR and CHAR to further isolate the effect of phase-based modeling versus second-order state input. On LAIR, where both methods can model the tasks, our approach produces more consistent results across seeds, while CONDOR occasionally mismodels segments (e.g., capricorn in Figure~\ref{fig:condor-pnpf-comp}). On CHAR, our method consistently outperforms the baseline, maintaining correct motion through multiple intersections, whereas CONDOR misses key segments (e.g., pulley, clover\_1 in Figure~\ref{fig:condor-pnpf-comp}). These results highlight the benefits of the phase-varying potential formulation. However, one limitation we observe is that our method tends to smooth out sharp corners, as seen in the capricorn task.

In addition, to empirically assess the stability, a finite-horizon rollout analysis across all tasks and seeds on LASA, LAIR, and CHAIR was added. The phase variable was discretized into 100 values and 1000 perturbed states was sampled per phase of the form $x^*(s) + u$, where $u$ was drawn uniformly from a local Euclidean ball. Evaluation was conducted for $\|u\|\leq 0.25r$, where $r$ denotes the radius of the data-supported region. We disabled the safeguard mechanism to evaluate the prevalence of undesired critical points. Across 100-step closed-loop rollouts, in over $99.999\%$ of sampled cases either both the phase variable and total potential decreased monotonically or the system reached the terminal region. Inspection of the rare failure cases indicates that they are associated with local gradient cancellation between the nominal and safety energy terms; enabling the safeguard resolves these cases, yielding a  $100\%$ success rate.

We evaluate reactivity to two types of perturbations: (i) obstacles, using the GShape task (LASA), and (ii) spatial disturbances, using the double\_loop task (LAIR). For obstacles, we demonstrate basic avoidance capabilities using simple repulsive potentials. For spatial disturbances, we additionally compare against CONDOR and Linear Parameter-Varying Dynamical Systems (LPVDS)~\footnote{We use the LPVDS implementation (\url{github.com/penn-figueroa-lab/lpvds}), extended to second-order systems with PQLF-based stability and slack-variable relaxation of the Lyapunov constraints. Clusters are defined from time-aligned trajectory segments, as EM-based clustering performed poorly.}. After the perturbation, the desired behavior is to resume the task smoothly without skipping or repeating segments. All methods restart from the disturbed position; baselines reset velocity while keeping the phase unchanged. This design introduces both temporal and spatial disturbances: velocity resets disrupt implicit progression, while the unchanged phase represents a temporal disturbance that the model must correct. Our method successfully recovers as expected, as it updates the phase in a progress-aware and locally consistent manner, whereas CONDOR and LPVDS tend to skip or repeat segments, relying on velocity to infer task progression. Moreover, CONDOR completes the task from an unnatural direction, and LPVDS deviates from the demonstrated path. This shows that our method remains robust to velocity disturbances, in contrast to second-order DS models, which may perform poorly with imprecise velocity observations.

\begin{figure*}[t]
\centering
\resizebox{0.75\linewidth}{!}{
\begin{minipage}[b]{0.479\linewidth}
  \centering
  \includegraphics[width=\linewidth]{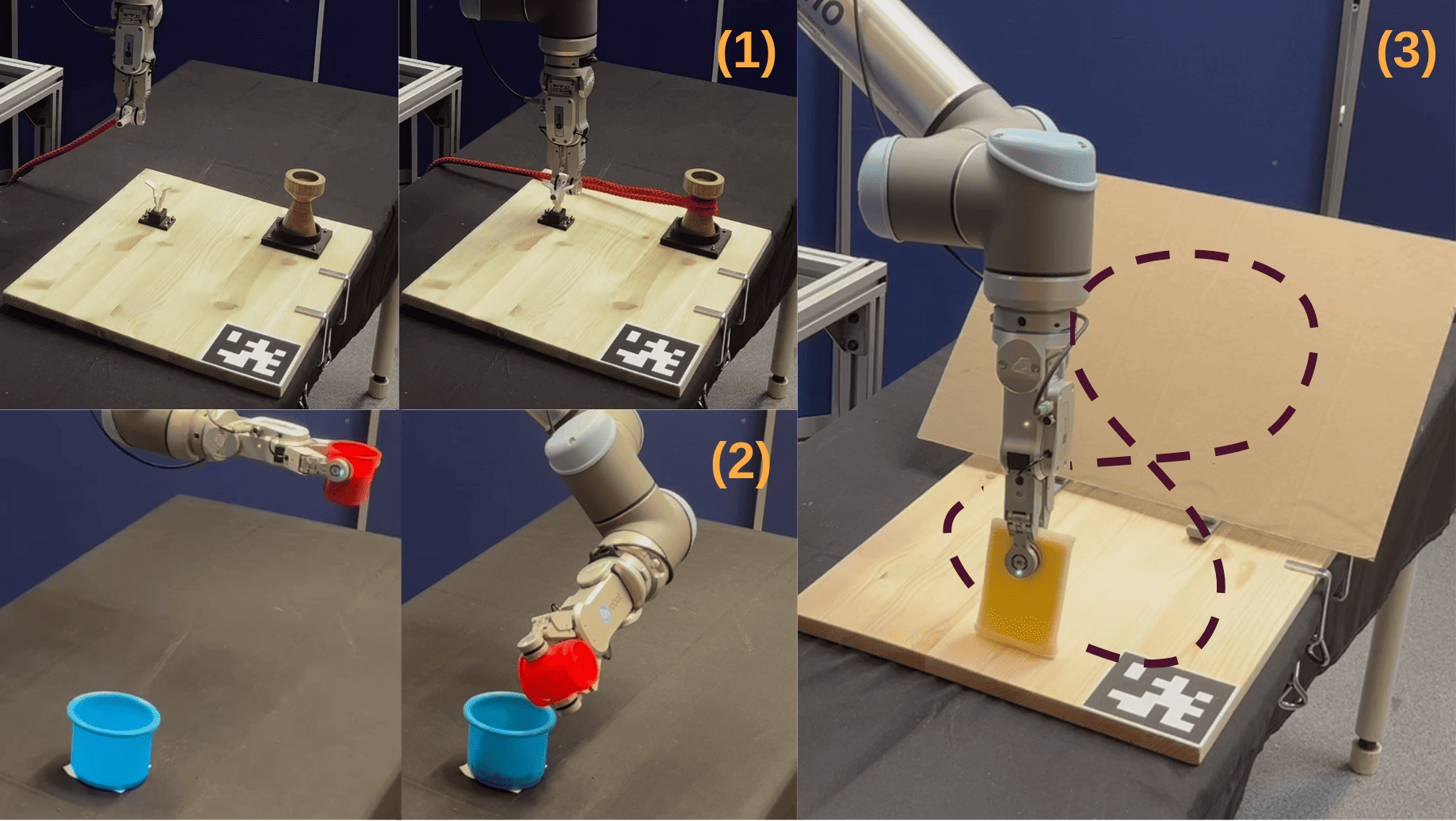}
\end{minipage}%
\hfill
\begin{minipage}[b]{0.49\linewidth}
  \centering
  \includegraphics[width=1\linewidth]{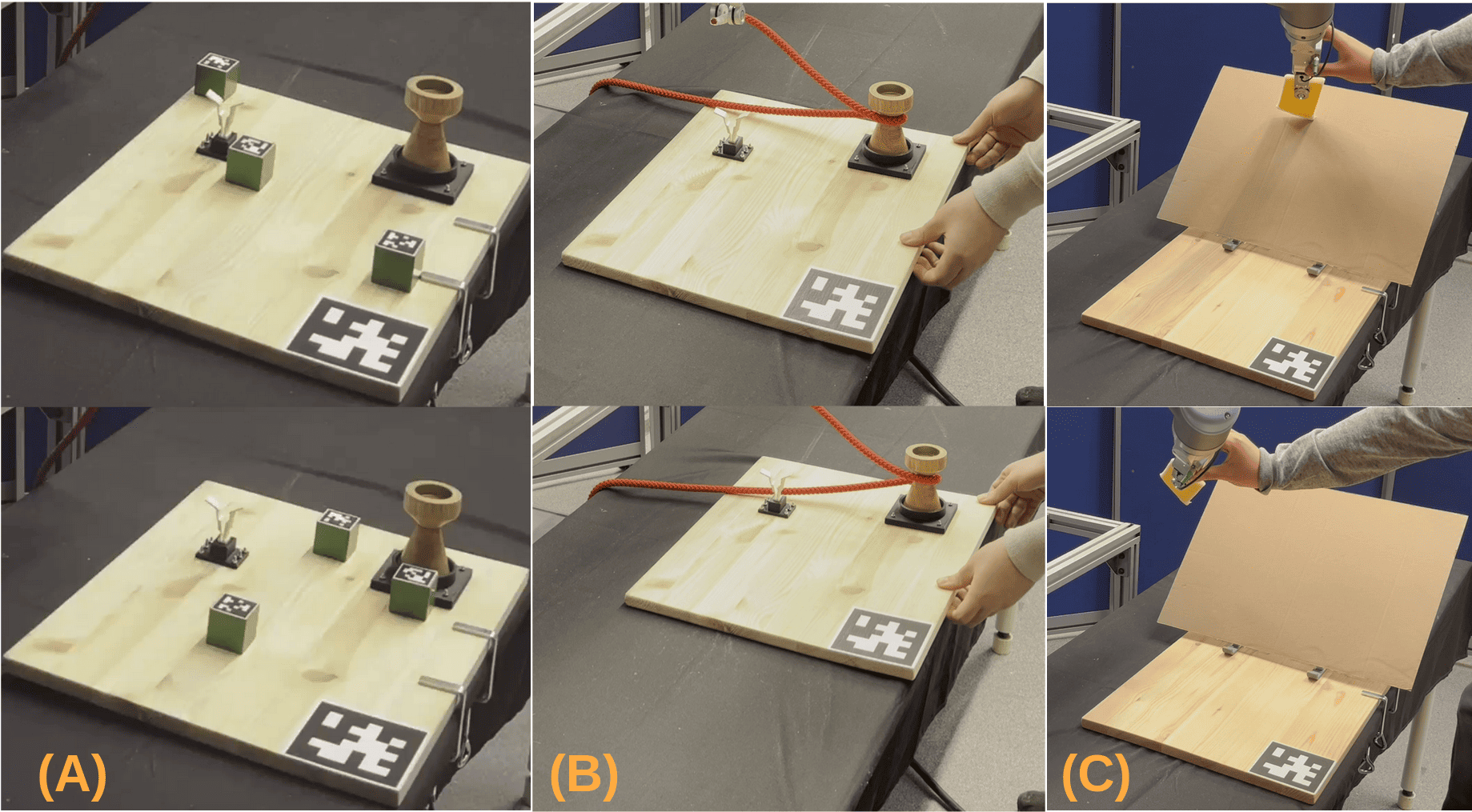}
\end{minipage}
}
\caption{Real robot experiments: Start and end configurations (left), with periodic motion in Task 3 indicated by a dashed line. (A) Obstacle variations with AprilTag-tracked cubes, (B) task frame shifts from moving the board mid-task, and (C) end-effector displacements from manual robot repositioning.}
\label{fig:real-robot}

\end{figure*}
\vspace{-1em}
\section{Real world experiments}

To showcase our method’s capabilities, we design and execute three real-world robotic tasks, illustrated in Figure~\ref{fig:real-robot}. The first is a knotting task, where the robot ties a rope around a cylinder for one full turn before inserting the rope's end into a designated socket. The second task is pouring, where the robot transfers a glass of beans to another container. The final task is a 3D wiping operation that features a periodic figure-eight motion. All experiments are carried out using a UR10 robot equipped with an RG2-FT gripper.

For model training, we collect demonstration data through kinesthetic teaching, recording six trajectories for each of the first two tasks. Given the periodic nature of the final task, fewer trajectories are used (only two). For the first and third tasks, we use positions and quaternions, while for the second task, we use joint angles, demonstrating that our method can handle various state representations. All training trajectories are time-aligned using DTW.

We first evaluate whether our model can generate appropriate motions when the robot starts from varying positions. For each task, we run tests from 10 different initial locations to assess the model's consistency and robustness. In the third task, to demonstrate periodic behavior, we initialize the phase variable $s$ as $2\times s_T$, producing two complete motion cycles. In all tests, the generated trajectories closely resemble the demonstrated ones, successfully complete their tasks, and exhibit the expected behavior. In these experiments, our framework is deployed online, operating at frequencies of up to 50 Hz.

Next, we perform an obstacle avoidance experiment in the first task to evaluate the reactivity of our method to environmental perturbations. In this setup, three cubes with side lengths of approximately 3.5 cm are placed within the workspace. Their positions relative to the task frame are tracked using AprilTags~\cite{apriltag}. These cubes are treated as static columns, and the obstacle energy function is computed based on their $x$ and $y$ coordinate~\footnote{Incorporating the $z$ coordinate trivializes the task, as the motion primarily occurs above the cubes.}. For simplicity, the robot’s end-effector is modeled as a point agent, with the obstacle radius adjusted accordingly. The experiment is repeated with two distinct obstacle configurations, both shown in Figure~\ref{fig:real-robot}A. In all cases, the robot avoids obstacles and successfully completes the tasks.

Finally, we conduct spatial disturbance experiments on the first and third tasks to further assess the responsiveness of our method to perturbations. Two types of disturbance are applied: (i) shifting the task frame and (ii) pausing the robot and manually displacing the end effector from its current pose. Examples of these disturbances are shown in Figure~\ref{fig:real-robot}B and C. The first type is applied only to the first task, as the task frame in the third task is fixed due to the occasional occlusion of the AprilTag used for tracking. Each task is tested with two distinct perturbation scenarios. In all cases, the robot successfully completes the tasks while demonstrating appropriate and adaptive behavior. Notably, no essential steps are skipped. For example, in the first task, when the robot is pushed backwards, causing the rope to untie from the cylinder, it responds correctly by re-tying the rope, completing the task as intended.
\vspace{-1em}
\section{Conclusion} 
\label{sec:conclusion}

This work introduces a novel LFD framework for reactive motion generation based on phase-varying neural potential functions. The proposed method models a wide range of tasks as dynamical systems governed by a closed-loop phase variable while preserving reactivity to external perturbations. Simulation results show that our approach outperforms baselines on the evaluated datasets in terms of average DTWD and FD metrics. The method effectively models dynamical systems that require time- or progress-based parameterization, improving upon the limitations of many existing DS-based reactive motion generation methods while maintaining reactive behavior. We demonstrate its effectiveness on real-robot control across three manipulation tasks, including both point-to-point and periodic motions. A current limitation is the tendency to oversmooth sharp corners due to ambiguities in the local energy formulation. Future work will refine the energy design to better preserve directional structure and improve robustness to demonstration artifacts, and incorporate visual and force feedback.

\section*{Acknowledgments}
We thank Bal\'azs Kulcs\'ar and Torsten Wik for their feedback on the stability analysis.
\vspace{-1em}
\bibliographystyle{IEEEtran}
\bibliography{IEEEabrv, references}

\end{document}

%% file: Results/short_results_one_column.tex
\begin{table}[t]
    \vspace{3mm}
\centering
\caption{Method performance across different datasets.}

\resizebox{0.95\linewidth}{!}{%
\begin{tabular}{c l c c c c}
\toprule
\textbf{Dataset} & \textbf{Method} & \textbf{DTWD} & \textbf{FD} & \textbf{FP Error} & \textbf{Accuracy} \\
\midrule

\multirow{3}{*}{\rotatebox{0}{LASA (cm)}} 
  & PNPF     & $\mathbf{1.81}$ & $\mathbf{3.38}$ & $\mathbf{0.02}$ & $\mathbf{1.00}$ \\
  & CONDOR  & $2.10$ & $4.14$ & $0.10$ & $\mathbf{1.00}$ \\
  & NODE  & $2.12$ & $4.33$ & $1.82$ & $0.76$\\
\midrule

\multirow{2}{*}{\rotatebox{0}{LAIR (mm)}} 
  & PNPF     & $\mathbf{4.82}$ & $\mathbf{11.75}$ & $\mathbf{5.04}$ & $\mathbf{1.00}$ \\
  & CONDOR  & $7.62$ & $17.43$ & $7.91$ & $0.96$ \\
\midrule

\multirow{2}{*}{\rotatebox{0}{CHAR (mm)}} 
  & PNPF     & $\mathbf{2.96}$ & $\mathbf{8.44}$ & $\mathbf{2.00}$ &  $\mathbf{1.00}$ \\
  & PNPF-50     & $3.16$ & $8.70$ & $2.02$ &  $\mathbf{1.00}$ \\
  & PNPF-H     & $3.35$ & $9.40$ & $2.01$ &  $\mathbf{1.00}$ \\
  & CONDOR  & $17.67$ & $45.56$ & $3.25$  & $\mathbf{1.00}$ \\
\midrule
\multirow{2}{*}{\rotatebox{0}{Char-Periodic (mm)}} 
  & PNPF     & $\mathbf{7.77}$ & $\mathbf{14.30}$ & NA & NA\\
  & Naive  & $10.70$ &  $24.28$ & NA & NA \\
\bottomrule
\end{tabular}

}

\resizebox{0.95\linewidth}{!}{%
\begin{tabular}{c l c c c c}
\toprule
\textbf{Dataset} & \textbf{Method} & \textbf{DTWD-P (cm)} & \textbf{DTWD-O (rad)} & \textbf{FP Error (cm)}& \textbf{F0 Error (rad)} \\
\midrule

\multirow{2}{*}{\rotatebox{0}{RoboTasks}} 
  & PNPF    & $\mathbf{2.19}$ & $\mathbf{0.08}$ & $\mathbf{0.25}$ & $0.17$ \\
  & NODE  & $2.62$ & $0.10$  & $1.51$ & $\mathbf{0.13}$\\
\bottomrule
\end{tabular}
}

\label{tab:dataset-comparison}
\end{table}

%% file: Results/Lair_results.tex
\begin{table}[t!]
\vspace{3mm}
\centering
\caption{LAIR and CHAR RESULTS}

\resizebox{1\linewidth}{!}{%
\begin{tabular}{|l|cc|cc|cc|cc|}
\hline
 & \multicolumn{2}{c|}{DTWD (mm)} & \multicolumn{2}{c|}{FD (mm)} & \multicolumn{2}{c|}{FP Error (m)} & \multicolumn{2}{c|}{Accuracy} \\
 & PNPF & CONDOR & PNPF & CONDOR & PNPF & CONDOR & PNPF & CONDOR \\
\hline
capricorn & $\mathbf{6.22}$ & $18.91$ & $\mathbf{18.33}$ & $49.99$ & $\mathbf{3.30}$ & $10.59$ & $\mathbf{1.00}$ & $0.67$ \\
double\_lag & $\mathbf{3.86}$ & $4.84$ & $\mathbf{10.35}$ & $10.99$ & $\mathbf{5.54}$ & $7.32$ & $\mathbf{1.00}$ & $\mathbf{1.00}$ \\
double\_loop & $\mathbf{5.81}$ & $10.09$ & $\mathbf{14.50}$ & $17.16$ & $\mathbf{8.07}$ & $8.63$ & $\mathbf{1.00}$ & $\mathbf{1.00}$ \\
e & $4.81$ & $\mathbf{4.54}$ & $\mathbf{11.15}$ & $12.36$ & $\mathbf{2.90}$ & $4.71$ & $\mathbf{1.00}$ & $\mathbf{1.00}$ \\
Lag & $4.15$ & $\mathbf{3.70}$ & $\mathbf{8.52}$ & $8.53$ & $\mathbf{4.07}$ & $5.38$ & $\mathbf{1.00}$ & $\mathbf{1.00}$ \\
phi & $\mathbf{4.05}$ & $5.79$ & $\mathbf{10.25}$ & $12.01$ & $\mathbf{6.51}$ & $11.32$ & $\mathbf{1.00}$ & $\mathbf{1.00}$ \\
triple\_loop & $\mathbf{4.43}$ & $6.94$ & $\mathbf{9.81}$ & $14.49$ & $\mathbf{3.31}$ & $6.55$ & $\mathbf{1.00}$ & $\mathbf{1.00}$ \\
two & $\mathbf{5.24}$ & $6.16$ & $\mathbf{11.10}$ & $13.87$ & $\mathbf{6.64}$ & $8.78$ & $\mathbf{1.00}$ & $\mathbf{1.00}$ \\
\hline
pulley & $\mathbf{3.35}$ & $15.71$ & $\mathbf{9.54}$ & $55.46$ & $\mathbf{1.23}$ & $2.81$ & $\mathbf{1.00}$ & $\mathbf{1.00}$ \\
eyes & $\mathbf{1.97}$ & $18.69$ & $\mathbf{5.87}$ & $46.76$ & $\mathbf{1.27}$ & $1.31$ & $\mathbf{1.00}$ & $\mathbf{1.00}$ \\
clover\_1 & $\mathbf{3.18}$ & $25.64$ & $\mathbf{9.35}$ & $64.88$ & $\mathbf{3.01}$ & $3.41$ & $\mathbf{1.00}$ & $\mathbf{1.00}$ \\
clover\_2 & $\mathbf{3.31}$ & $15.68$ & $\mathbf{9.77}$ & $23.18$ & $\mathbf{3.62}$ & $5.57$ & $\mathbf{1.00}$ & $\mathbf{1.00}$ \\
double\_knot & $\mathbf{3.00}$ & $12.62$ & $\mathbf{7.65}$ & $37.53$ & $\mathbf{0.88}$ & $3.15$ & $\mathbf{1.00}$ & $\mathbf{1.00}$ \\
\hline
\end{tabular}
}
\label{Lair-results}
\end{table}

%% file: references.bib
@article{argall2009survey,
  title={A survey of robot learning from demonstration},
  author={Argall, Brenna D and Chernova, Sonia and Veloso, Manuela and Browning, Brett},
  journal={Robot. Auton. Syst.},
  volume={57},
  number={5},
  pages={469--483},
  year={2009},
  publisher={Elsevier}
}

@article{schaal1996learning,
  title={Learning from demonstration},
  author={Schaal, Stefan},
  journal={Adv. Neural Inf. Process. Syst.},
  volume={9},
  year={1996}
}

@inproceedings{medina2017learning,
  title={Learning stable task sequences from demonstration with linear parameter varying systems and hidden Markov models},
  author={Medina, Jose R and Billard, Aude},
  booktitle={Conf. Robot. Learn. (CoRL)},
  pages={175--184},
  year={2017},
}

@inproceedings{loshchilovdecoupled,
title={Decoupled Weight Decay Regularization},
author={Ilya Loshchilov and Frank Hutter},
booktitle={Int. Conf. Learn. Represent. (ICLR)},
year={2019},
}

@article{dmp,
  title={Dynamical movement primitives: learning attractor models for motor behaviors},
  author={Ijspeert, Auke Jan and Nakanishi, Jun and Hoffmann, Heiko and Pastor, Peter and Schaal, Stefan},
  journal={Neural computation},
  volume={25},
  number={2},
  pages={328--373},
  year={2013},
  publisher={MIT Press One Rogers Street, Cambridge, MA 02142-1209, USA journals-info~…}
}

@article{promp,
  title={Probabilistic movement primitives},
  author={Paraschos, Alexandros and Daniel, Christian and Peters, Jan R and Neumann, Gerhard},
  journal={Adv. Neural Inf. Process. Syst.},
  volume={26},
  year={2013}
}

@inproceedings{cnmp,
  title={Conditional Neural Movement Primitives.},
  author={Seker, Muhammet Yunus and Imre, Mert and Piater, Justus H and Ugur, Emre},
  booktitle={Robot.: Sci. Syst. (RSS)},
  volume={10},
  year={2019}
}

@article{li2023prodmp,
  title={Prodmp: A unified perspective on dynamic and probabilistic movement primitives},
  author={Li, Ge and Jin, Zeqi and Volpp, Michael and Otto, Fabian and Lioutikov, Rudolf and Neumann, Gerhard},
  journal={IEEE Robot. Autom. Lett.},
  volume={8},
  number={4},
  pages={2325--2332},
  year={2023},
}

@article{bahl2020neural,
  title={Neural dynamic policies for end-to-end sensorimotor learning},
  author={Bahl, Shikhar and Mukadam, Mustafa and Gupta, Abhinav and Pathak, Deepak},
  journal={Adv. Neural Inf. Process. Syst.},
  volume={33},
  pages={5058--5069},
  year={2020}
}

@article{yildirim2024conditional,
  title={Conditional neural expert processes for learning movement primitives from demonstration},
  author={Yildirim, Yigit and Ugur, Emre},
  journal={IEEE Robot. Autom. Lett.},
  year={2024},
}

@article{khansari2011learning,
  title={Learning stable nonlinear dynamical systems with gaussian mixture models},
  author={Khansari-Zadeh, S Mohammad and Billard, Aude},
  journal={IEEE Trans. Robot.},
  volume={27},
  number={5},
  pages={943--957},
  year={2011},
}

@article{khadivar2021learning,
  title={Learning dynamical systems with bifurcations},
  author={Khadivar, Farshad and Lauzana, Ilaria and Billard, Aude},
  journal={Robot. Auton. Syst.},
  volume={136},
  pages={103700},
  year={2021},
  publisher={Elsevier}
}

@book{billard2022learning,
  title={Learning for adaptive and reactive robot control: a dynamical systems approach},
  author={Billard, Aude and Mirrazavi, Sina and Figueroa, Nadia},
  year={2022},
  publisher={Mit Press}
}

@inproceedings{figueroa2018physically,
  title={A physically-consistent bayesian non-parametric mixture model for dynamical system learning},
  author={Figueroa, Nadia and Billard, Aude},
  booktitle={Conf. Robot. Learn. (CoRL)},
  pages={927--946},
  year={2018},
}

@inproceedings{apriltag,
  author = {Wang, John and Olson, Edwin},
  booktitle={IEEE/RSJ Int. Conf. Intell. Robot. Syst. (IROS)},
  doi = {10.1109/IROS.2016.7759617},
  isbn = {978-1-5090-3762-9},
  pages = {4193--4198},
  title = {{AprilTag 2: Efficient and robust fiducial detection}},
  year = {2016}
}

@inproceedings{figueroa2020dynamical,
  title={A dynamical system approach for adaptive grasping, navigation and co-manipulation with humanoid robots},
  author={Figueroa, Nadia and Faraji, Salman and Koptev, Mikhail and Billard, Aude},
  booktitle={IEEE Int. Conf. Robot. Autom. (ICRA)},
  pages={7676--7682},
  year={2020},
}

@inproceedings{grasp_df1,
  title={Neural Grasp Distance Fields for Robot Manipulation},
  author={Weng, Thomas and Held, David and Meier, Franziska and Mukadam, Mustafa},
  booktitle={IEEE Int. Conf. Robot. Autom. (ICRA)},
  pages={1814--1821},
  year={2023},
}

@article{irshad2024neural,
  title={Neural Fields in Robotics: A Survey},
  author={Irshad, Muhammad Zubair and others},
  journal={arXiv preprint arXiv:2410.20220},
  year={2024}
}

@article{tekden2023grasp,
  title={Grasp transfer based on self-aligning implicit representations of local surfaces},
  author={Tekden, Ahmet and Deisenroth, Marc Peter and Bekiroglu, Yasemin},
  journal={IEEE Robot. Autom. Lett.},
  volume={8},
  number={10},
  pages={6315--6322},
  year={2023},
}

@article{figueroa2022locally,
  title={Locally active globally stable dynamical systems: Theory, learning, and experiments},
  author={Figueroa, Nadia and Billard, Aude},
  journal={Int. J. Robot. Res.},
  volume={41},
  number={3},
  pages={312--347},
  year={2022},
  publisher={SAGE Publications Sage UK: London, England}
}

@inproceedings{deepsdf,
  title={Deepsdf: Learning continuous signed distance functions for shape representation},
  author={Park, Jeong Joon and Florence, Peter and Straub, Julian and Newcombe, Richard and Lovegrove, Steven},
  booktitle={IEEE/CVF Conf. Comput. Vis. Pattern Recognit. (CVPR)},
  pages={165--174},
  year={2019}
}

@inproceedings{tekden2023neural,
  title={Neural field movement primitives for joint modelling of scenes and motions},
  author={Tekden, Ahmet and Deisenroth, Marc Peter and Bekiroglu, Yasemin},
  booktitle={IEEE/RSJ Int. Conf. Intell. Robot. Syst. (IROS)},
  pages={3648--3655},
  year={2023},
}

@inproceedings{hypernet,
title={Hypernetworks},
author={Ha, David and Dai, Andrew M and Le, Quoc V},
  booktitle={Int. Conf. Learn. Represent. (ICLR)},
year={2017}
}

@incollection{dtw,
  author={M{\"u}ller, Meinard},
  title={Dynamic time warping},
  booktitle={Information Retrieval for Music and Motion},
  pages={69--84},
  year={2007},  
publisher={Springer}

}

@article{explog_map,
  title={Practical parameterization of rotations using the exponential map},
  author={Grassia, F Sebastian},
  journal={J. Graph. Tools},
volume={3},
  number={3},
  pages={29--48},
  year={1998},
  publisher={Taylor \& Francis}
}

@inproceedings{ude_quat,
  title={Orientation in cartesian space dynamic movement primitives},
  author={Ude, Ale{\v{s}} and Nemec, Bojan and Petri{\'c}, Tadej and Morimoto, Jun},
  booktitle={IEEE Int. Conf. Robot. Autom. (ICRA)},
  pages={2997--3004},
  year={2014},
}

@article{khatib1986real,
  title={Real-time obstacle avoidance for manipulators and mobile robots},
  author={Khatib, Oussama},
  journal={Int. J. Robot. Res.},
  volume={5},
  number={1},
  pages={90--98},
  year={1986},
  publisher={Sage Publications Sage CA: Thousand Oaks, CA}
}

@article{perez2023stable,
  title={Stable motion primitives via imitation and contrastive learning},
  author={P{\'e}rez-Dattari, Rodrigo and Kober, Jens},
  journal={IEEE Trans. Robot.},
  year={2023},
}

@inproceedings{node-clf-cbm,
  title={Learning Complex Motion Plans using Neural ODEs with Safety and Stability Guarantees},
  author={Nawaz, Farhad and Li, Tianyu and Matni, Nikolai and Figueroa, Nadia},
  booktitle={IEEE Int. Conf. Robot. Autom. (ICRA)},
  pages={17216--17222},
  year={2024},
}

@article{eiter1994computing,
  title={Computing discrete Fr{\'e}chet distance},
  author={Eiter, Thomas and Mannila, Heikki},
  year={1994},
  journal={Technical Report CD-TR 94/64},
  institution={Christian Doppler Laboratory for Expert Systems, Vienna University of Technology}
}

@article{clfd,
  title={Continual learning from demonstration of robotics skills},
  author={Auddy, Sayantan and Hollenstein, Jakob and Saveriano, Matteo and Rodr{\'\i}guez-S{\'a}nchez, Antonio and Piater, Justus},
  journal={Robot. Auton. Syst.},
  volume={165},
  pages={104427},
  year={2023},
  publisher={Elsevier}
}

@misc{clfd-snode,
      title={Scalable and Efficient Continual Learning from Demonstration via a Hypernetwork-generated Stable Dynamics Model}, 
      author={Sayantan Auddy and Jakob Hollenstein and Matteo Saveriano and Antonio Rodríguez-Sánchez and Justus Piater},
      year={2024},
      eprint={2311.03600},
      archivePrefix={arXiv},
      primaryClass={cs.RO}
}

@article{nerf_survey,
  journal={Comput. Graph. Forum},
   title = {Neural Fields in Visual Computing and Beyond},
    author = {Xie, Yiheng and others},
    year = {2022},
    publisher = {The Eurographics Association and John Wiley & Sons Ltd.},
    ISSN = {1467-8659},
    DOI = {10.1111/cgf.14505}
}

@article{tancik2020fourier,
  title={Fourier features let networks learn high frequency functions in low dimensional domains},
  author={Tancik, Matthew and others}, 
  journal={Adv. Neural Inf. Process. Syst.},
  volume={33},
  pages={7537--7547},
  year={2020}
}

@article{siren,
  title={Implicit neural representations with periodic activation functions},
  author={Sitzmann, Vincent and Martel, Julien and Bergman, Alexander and Lindell, David and Wetzstein, Gordon},
  journal={Adv. Neural Inf. Process. Syst.},
  volume={33},
  pages={7462--7473},
  year={2020}
}

@article{khansari2017learning,
  title={Learning potential functions from human demonstrations with encapsulated dynamic and compliant behaviors},
  author={Khansari-Zadeh, Seyed Mohammad and Khatib, Oussama},
  journal={Auton. Robots},
  volume={41},
  pages={45--69},
  year={2017},
  publisher={Springer}
}

@article{totsila2025sensorimotor,
  title={Sensorimotor Learning with Stability Guarantees via Autonomous Neural Dynamic Policies},
  author={Totsila, Dionis and Chatzilygeroudis, Konstantinos and Modugno, Valerio and Hadjivelichkov, Denis and Kanoulas, Dimitrios},
  journal={IEEE Robot. Autom. Lett.},
  year={2025},
}

@inproceedings{chen2025implicit,
  title={Implicit Articulated Robot Morphology Modeling with Configuration Space Neural Signed Distance Functions},
  author={Chen, Yiting and Gao, Xiao and Yao, Kunpeng and Niederhauser, Lo{\"\i}c and Bekiroglu, Yasemin and Billard, Aude},
  booktitle={IEEE Int. Conf. Robot. Autom. (ICRA)},
  pages={4558--4564},
  year={2025},
}

@ARTICLE{ames2017cbf,
  author={Ames, Aaron D. and Xu, Xiangru and Grizzle, Jessy W. and Tabuada, Paulo},
  journal={IEEE Trans. Autom. Control},
  title={Control Barrier Function Based Quadratic Programs for Safety Critical Systems}, 
  year={2017},
  volume={62},
  number={8},
  pages={3861-3876},
}

@inproceedings{saveriano2019learning,
  title={Learning barrier functions for constrained motion planning with dynamical systems},
  author={Saveriano, Matteo and Lee, Dongheui},
  booktitle={IEEE/RSJ Int. Conf. Intell. Robot. Syst. (IROS)},
  pages={112--119},
  year={2019},
}

@inproceedings{rana2020learning,
  title={Learning reactive motion policies in multiple task spaces from human demonstrations},
  author={Rana, M Asif and others},
  booktitle={Conf. Robot. Learn. (CoRL)},
  pages={1457--1468},
  year={2020},
}
